%% file: ms.tex
\documentclass[11pt]{article} 

\usepackage{graphicx}
\usepackage{tabularx}
\usepackage{booktabs}
\usepackage[T1]{fontenc}
\usepackage{lmodern}
\usepackage[flushleft]{threeparttable}
\usepackage[hidelinks]{hyperref} %
\hypersetup{hidelinks}
\usepackage{multirow}
\usepackage{longtable}
\usepackage[margin=3cm]{geometry} %
\usepackage{amsmath}
\usepackage{amssymb}
\usepackage{authblk}
\usepackage{caption}

\title{K-complex Detection Using Fourier Spectrum Analysis In EEG}
\author[1]{Alexey Protopopov}
\affil[1]{Joint Stock "Research and production company "Kryptonite" \authorcr
E-mail: a.protopopov@kryptonite.ru}
\date{}
\begin{document}
    \captionsetup[figure]{labelformat={default},labelsep=period,name={Figure}}
    \captionsetup[table]{labelformat={default},labelsep=period,name={Table}}

    \maketitle

    \begin{abstract}
        K-complexes are an important marker of brain activity and are used both in clinical practice to perform sleep scoring, and in research. However, due to the size of electroencephalography (EEG) records, as well as the subjective nature of K-complex detection performed by somnologists, it is reasonable to automate K-complex detection. Previous works in this field of research have relied on the values of true positive rate and false positive rate to quantify the effectiveness of proposed methods, however this set of metrics may be misleading. The objective of the present research is to find a more accurate set of metrics and use them to develop a new method of K-complex detection, which would not rely on neural networks. Thus, the present article proposes two new methods for K-complex detection based on the fast Fourier transform. The results achieved demonstrated that the proposed methods offered a quality of K-complex detection that is either similar or superior to the quality of the methods demonstrated in previous works, including the methods employing neural networks, while requiring less computational power, meaning that K-complex detection does not require the use of neural networks. The proposed methods were evaluated using a new set of metrics, which is more representative of the quality of K-complex detection.
    \end{abstract}


    \input{full}

\end{document}

%% file: full.tex
\section{Introduction}\label{introduction}

Electroencephalography (EEG) is a very popular method used in both clinical and scientific studies of brain activity. It has seen significant development over the course of the past half-century and one of the current goals of EEG studies is automated sleep analysis. Considering the size of a typical EEG signal recorded overnight, it is quite obvious that even partial automation of the annotation process would be of great benefit to both clinical staff and scientists using EEG in their studies. The detection of K-complexes, an important marker of brain activity, is of particular interest, and there have been numerous works devoted to this problem. The methods proposed in such works vary, for example the authors of \cite{bankman} used neural networks trained on raw signals and features extracted from them. This work is widely cited and may be considered to be a gold standard of K-complex detection. Another method, proposed in \cite{kam}, uses two competing hidden Markov models, one trained to detect K-complexes and another trained to detect the signal without a K-complex. Finally, the algorithm presented in \cite{yucelbas} uses three different mathematical methods to identify K-complexes without the use of neural networks. Most of the existing works on the subject use the same set of parameters to estimate the quality of the proposed detection methods: true positive rate (TPR) and false positive rate (FPR).
In the course of the present work, we have developed and tested two new methods of K-complex detection that are not based on the use of neural networks: the K-complex Band Power (KBP) and the Harmonic Coordinate Matching (HCM) algorithms. Furthermore, we propose a new set of metrics for estimating the effectiveness of such methods, which we believe is more appropriate than the existing set of metrics.

\section{Metrics}\label{metrics}

As it was mentioned before, previous works on the subject of K-complex detection used the combination of TPR and FPR to demonstrate the quality of presented methods. These parameters are calculated using the following formulae:

\begin{equation}
    TPR = \ \frac{TP}{TP + FN}
    \label{eq:1}
\end{equation}
\begin{equation}
    FPR = \ \frac{FP}{FP + TN}
    \label{eq:2}
\end{equation}

Here TP, FP, TN, and FN are the numbers of true positives, false positives, true negatives, and false negatives respectively. Ideally, TPR should tend to 100\%, while FPR should tend to 0\% for optimum detection quality. A registration is considered to be a TP when it falls within a certain time window around an annotation made by a human, otherwise it is considered to be a FP. Similarly, a FN occurs when the algorithm does not register a K-complex near a human annotation. The method of determining FNs is somewhat more complex: the signal is split into intervals 1 second long, and the ones that contain neither human nor algorithm annotations are considered to be TNs.

However, due to their nature, the total number of K-complexes within an EEG record will be an order of magnitude less than the number of 1 second intervals without them, meaning that in the abovementioned formulae TN will be an order of magnitude greater than every other variable. Indeed, let’s assume that a 100-second-long record has 10 K-complexes, and that an algorithm has made 20 registrations: 10 of them true and 10 – false. Simple calculations show that in this case:

\begin{equation}
    TPR = \ \frac{10}{10 + 0} = \ 100\%
    \label{eq:3}
\end{equation}
\begin{equation}
    FPR = \ \frac{10}{10 + 80} \approx \ 11\%
    \label{eq:4}
\end{equation}

These are good numbers compared to to the results published in existing papers, but they conceal the fact that half of the registrations made by the algorithm are incorrect. Furthermore, the value of FPR depends on the frequency of occurrence of K-complexes in the signal, meaning that this value will be affected by the dataset chosen for the experiment.

Thus, we deem it reasonable to propose an alternate set of metrics that would be more representative of the quality of K-complex detection. This new metric is composed of the TPR and the positive predictive value (PPV), defined as follows:

\begin{equation}
    PPV = \ \frac{TP}{TP + FP}
    \label{eq:5}
\end{equation}

which would tend to 100\% in ideal conditions. The value of the TPR reflects how many of the K-complexes labeled by the human were detected by the algorithm, while the PPV reflects how many of the registrations matched the human annotations. This set of metrics does not use FN, which means that it is not affected by the rate at which the K-complex appears in the EEG record. In case of the aforementioned example, PPV would be equal to 50\%, which offers a much clearer representation of the K-complex detection quality and does not depend on the frequency of occurrence of K-complexes in the signal.

\section{Materials}\label{materials}

All EEG records used in the present article were acquired and annotated at the Institute of Higher Nervous Activity and Neurophysiology of the Russian Academy of Sciences. The records were made over 20~channels with a sampling frequency of 250~Hz, were taken from 5~different individuals, and had a total duration of approximately 10~hours.

\section{Methods}\label{methods}

\subsection{KBP algorithm}\label{meth_kbp_algorithm}

The KBP algorithm uses a sliding window which moves along the time axis of each channel. For each window, a set of five values is calculated:

\begin{enumerate}
    \item \emph{P\textsubscript{1}} – total power within the span of {[}0.0;~3.5{]}~Hz,
    \item \emph{P\textsubscript{2}} – total power within the span of {[}1.0;~4.5{]}~Hz,
    \item \emph{P\textsubscript{3}} – total power within the span of {[}2.0;~5.5{]}~Hz,
    \item \emph{P\textsubscript{4}} – total power within the span of {[}3.0;~6.5{]}~Hz,
    \item \emph{S} – amplitude difference between the absolute maximum and the following local minimum within the window.
\end{enumerate}

\begin{figure}[t]
    \centering
    \includegraphics[scale=0.7]{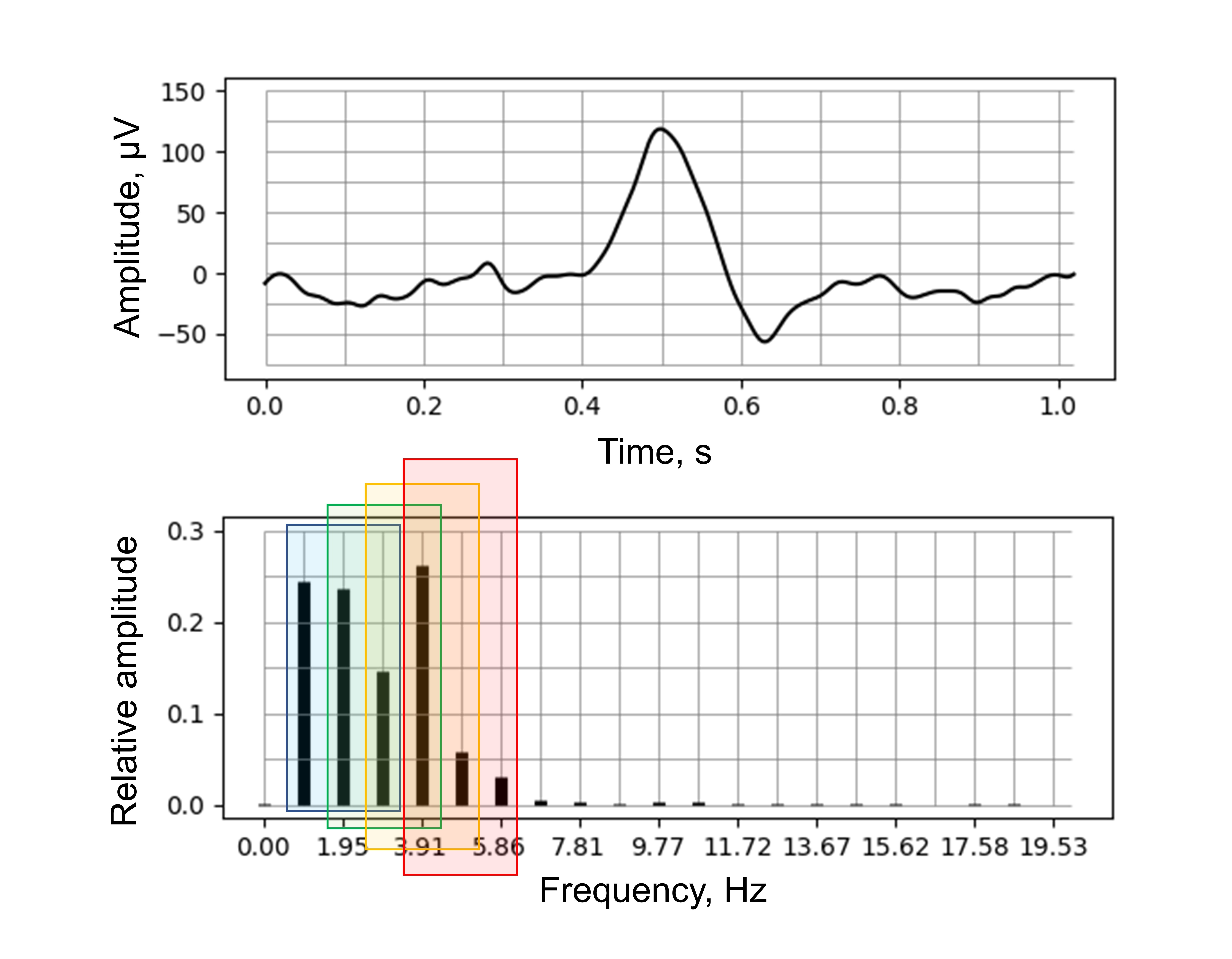}
    \caption{A typical K-complex with its power spectrum. The frequency bands analyzed by the algorithm are marked with blue, green, orange and red, respectively.}
    \label{fig:1}
\end{figure}

The values of \emph{P\textsubscript{i}} are calculated by applying the fast Fourier transform (FFT) to the signal within the window and calculating the sum of the powers of each harmonic within the specified frequency range. This is illustrated by Figure~\ref{fig:1}.

\begin{figure}[t]
    \centering
    \includegraphics[scale=0.7]{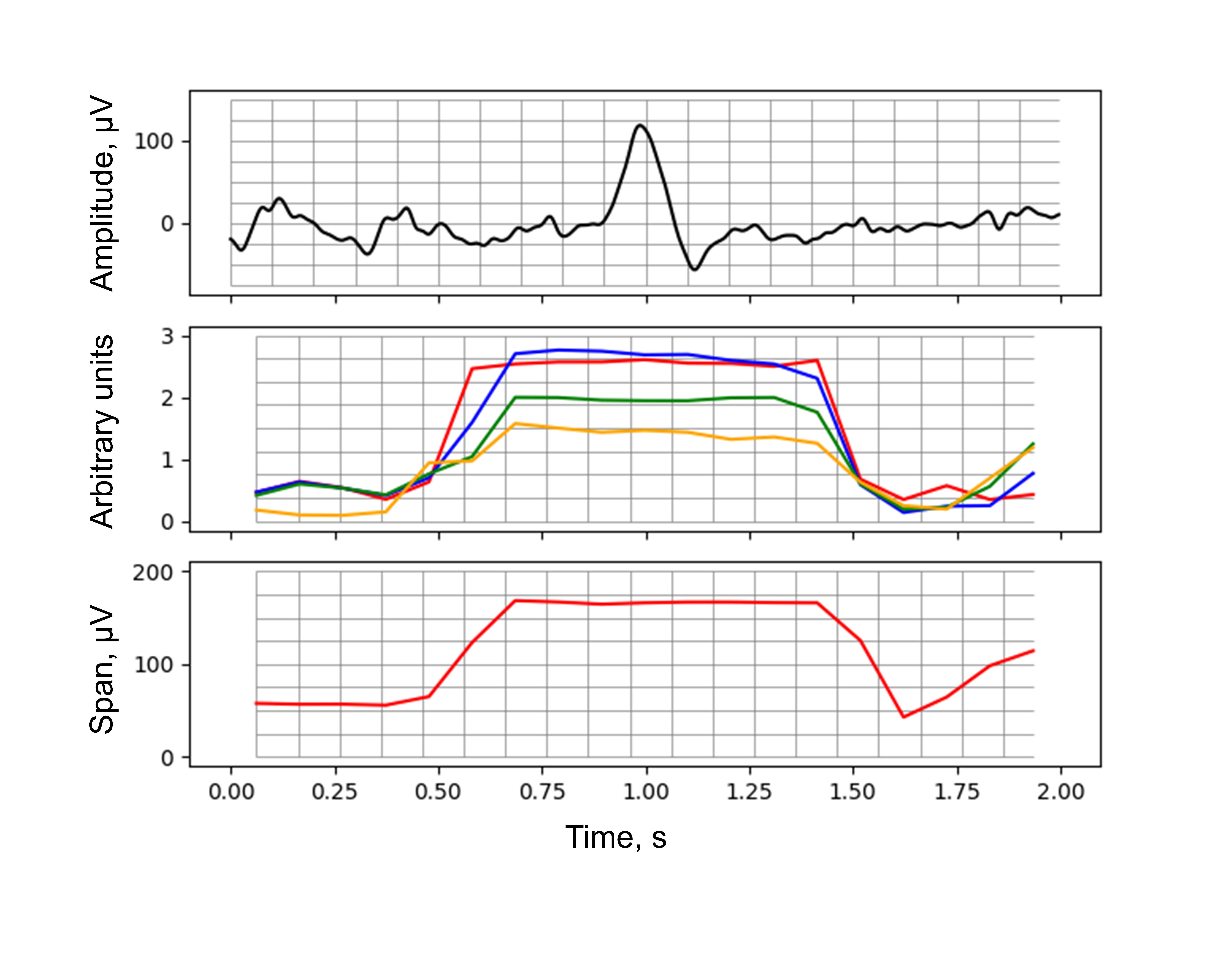}
    \caption{Top: the same K-complex as in Fig.~\ref{fig:1}. Middle: graphs of \emph{P\textsubscript{1-4}} in red, blue, green, and orange, respectively. Bottom: graph of \emph{S}.}
    \label{fig:2}
\end{figure}

The value of \emph{S} is calculated by taking the difference between the amplitude of the absolute maximum of the signal within the window, and the amplitude of the first minimum following it. This may seem counterintuitive at first, since according to \cite{aasm}, the K-complex is a “well-delineated, negative, sharp wave immediately followed by a positive component”. However, the polarity of the K-complex is ambiguous, meaning that depending on the conditions of the recording, it may begin with a sharp positive wave, rather than negative. This means that by calculating \emph{S} twice – once for the original signal and once for an inverted one, and then taking the greater value of the two, we essentially calculate the absolute value of the amplitude span of the K-complex if it is present in the window. However, this simple approach is complicated by the fact that EEG recordings often contain components of relatively high frequencies (over 10~Hz), which are of no interest to us, but may distort the shape of the K-complex. The algorithm compensates for that by applying the FFT to the signal within the window, discarding the coefficients corresponding to the frequencies above 10~Hz, reconstructing the signal using the remaining coefficients and using the smooth reconstructed signal to calculate \emph{S}. The way this and the other feature values behave around the K-complex is shown in Figure~\ref{fig:2}.

\begin{figure}[t]
    \centering
    \includegraphics[scale=0.5]{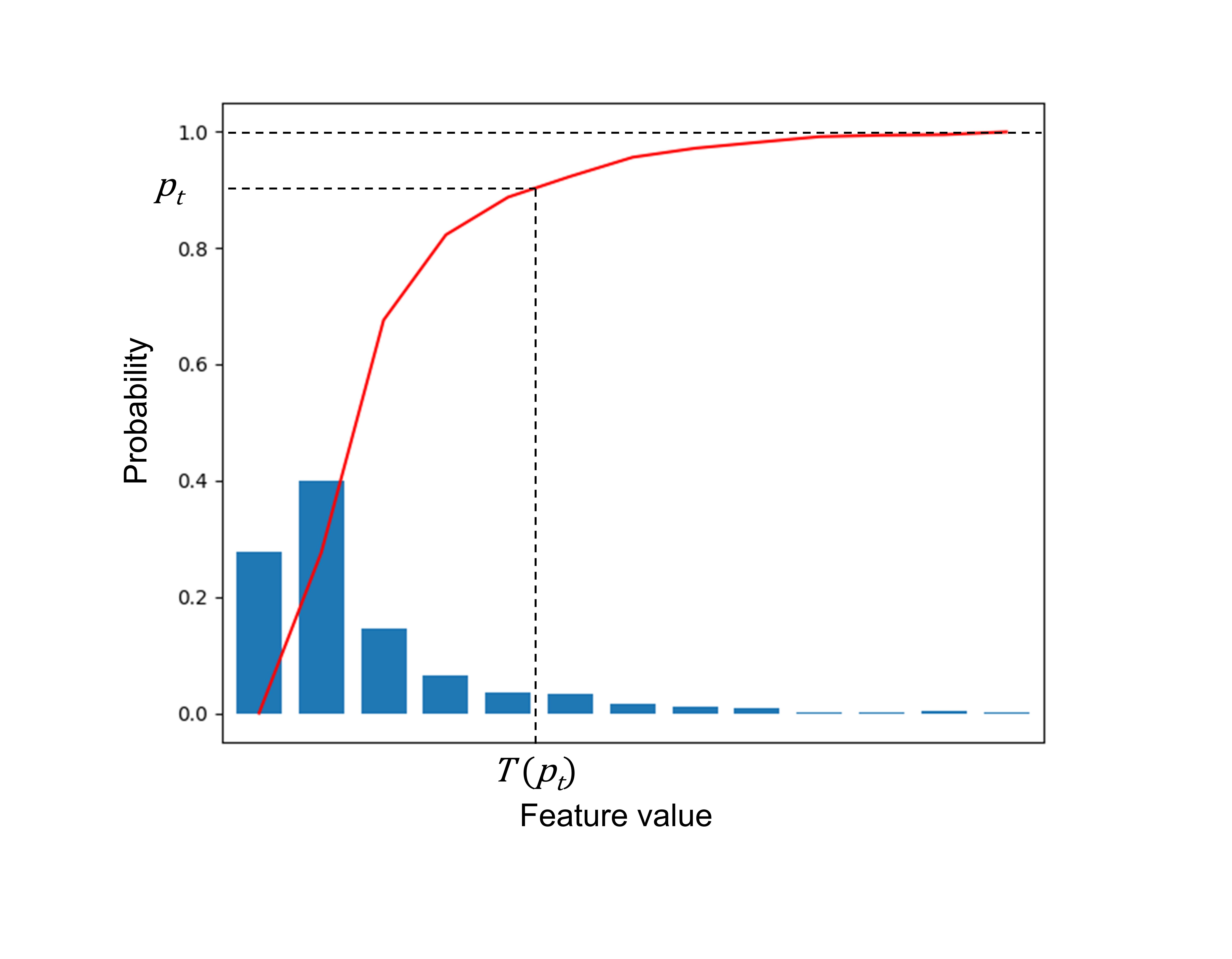}
    \caption{The histogram of feature values within a signal (blue) together with the integral probability curve (red).}
    \label{fig:3}
\end{figure}

Once all five values are calculated, they are compared against the corresponding thresholds. Thresholds are calculated separately for each of the five features used. First, the values of the feature are arranged into a histogram, which is used to calculate the integral probability curve. Each of the features has a tuning coefficient associated with it, labelled \emph{p\textsubscript{t}}, which is used to calculate the threshold \emph{T}(\emph{p\textsubscript{t}}) as shown in Figure~\ref{fig:3}. In essence this coefficient is the probability of encountering a feature value less than the threshold.

If the value of a specific feature is greater than the threshold, the corresponding window has a logic value \emph{l\textsubscript{i}}(\emph{f\textsubscript{i}}) set to 1, where \emph{f\textsubscript{i}} is the value of one of the features within the window and \emph{i} is the window index. Otherwise, this value is set to 0. These logic values are combined according to the formula:

\begin{equation}
    (l_i{P_{1,i}})\vee \ (l_i{P_{2,i}})\vee \ (l_i{P_{3,i}})\vee \ (l_i{P_{4,i}})\vee \ (l_i{S_i}) = \ L_i
    \label{eq:6}
\end{equation}

The resulting value \emph{L\textsubscript{i}} determines whether the algorithm found a K-complex within the window, with 1 meaning “yes” and 0 meaning “no”.

Finally, the algorithm takes the results from all the channels of the recording and sums them together. This means that every window is now labeled with an integer ranging from 0 to \emph{M}, where \emph{M} is the number of channels in the record. If this value is greater than a threshold \emph{V\textsubscript{t}}, then the window is marked as containing a K-complex, otherwise it is left unmarked.

\subsection{HCM algorithm}\label{meth_hcm_algorithm}

\begin{figure}[t]
    \centering
    \includegraphics[scale=0.5]{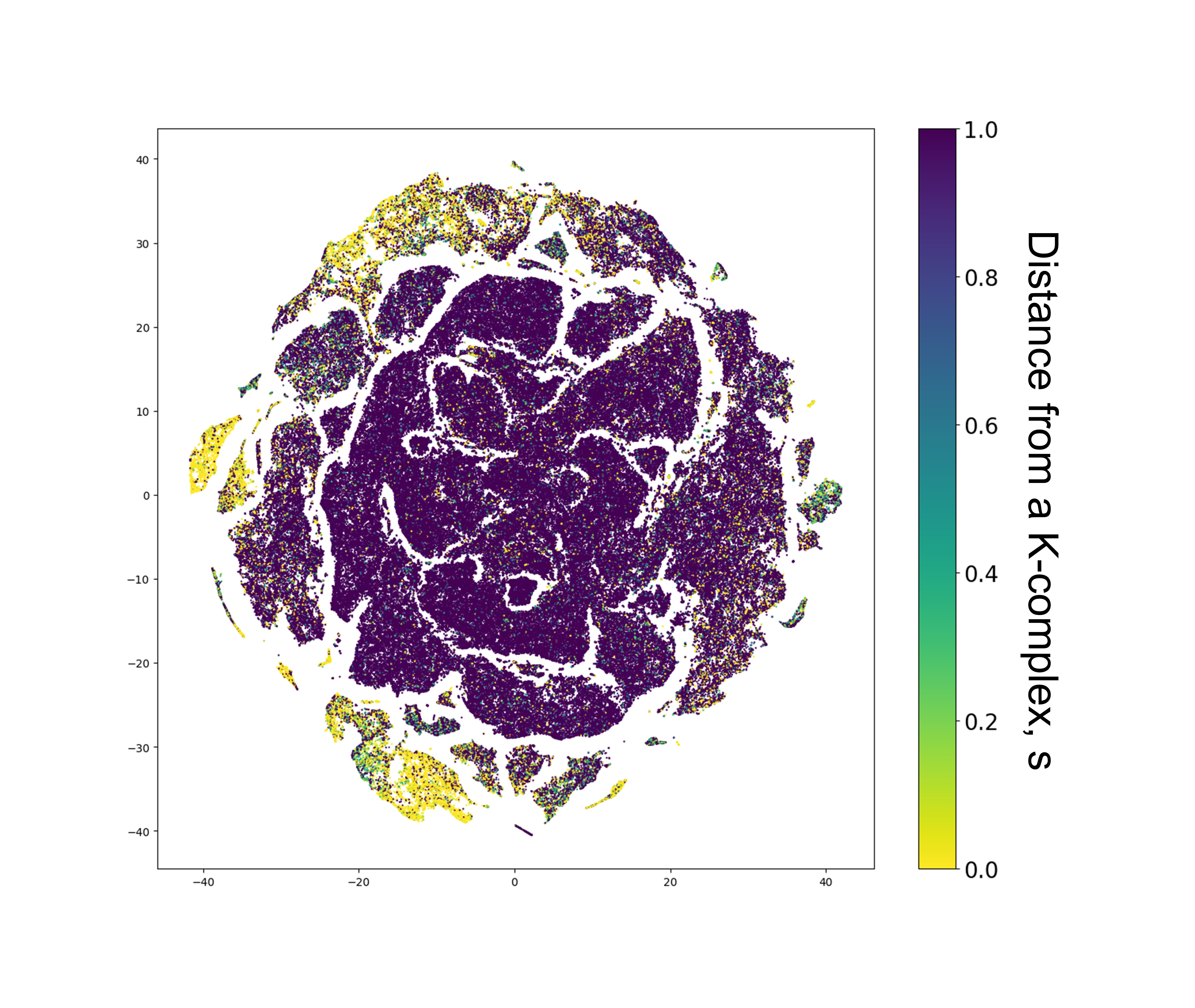}
    \caption{t-SNE representation of the distribution of windows in a harmonic space. Harmonics used: 0~–~9.}
    \label{fig:4}
\end{figure}

The Harmonic Coordinate Matching algorithm is based on the assumption that the spectral makeup of the signal in the vicinity of a K-complex is different from that of the signal elsewhere. The spectra are calculated for each window the same way as they were calculated in the KBP algorithm. Thus, taking the powers in each harmonic as a set of vector coordinates in an \emph{N}-dimensional space, where \emph{N} is the number of harmonics used, we represent each window as a point in the \emph{N}-dimensional harmonic space. Based on the aforementioned assumption we expect that the points corresponding to the windows in the vicinity of the K-complex to be separated from the other points. To check this theory, we take the first 10~harmonics, plot the points in the 10-dimensional space and visualize the result using the t-SNE algorithm (Figure~\ref{fig:4}).

It is clear from the figure above that the points tend to form clusters based on their proximity to K-complexes. By sequentially eliminating the harmonics from the experiment, we have determined, that the harmonics essential to window clusterization correspond to approximately 1, 2, 3, 4 and 7~Hz.

\begin{figure}[t]
    \centering
    \includegraphics[scale=0.5]{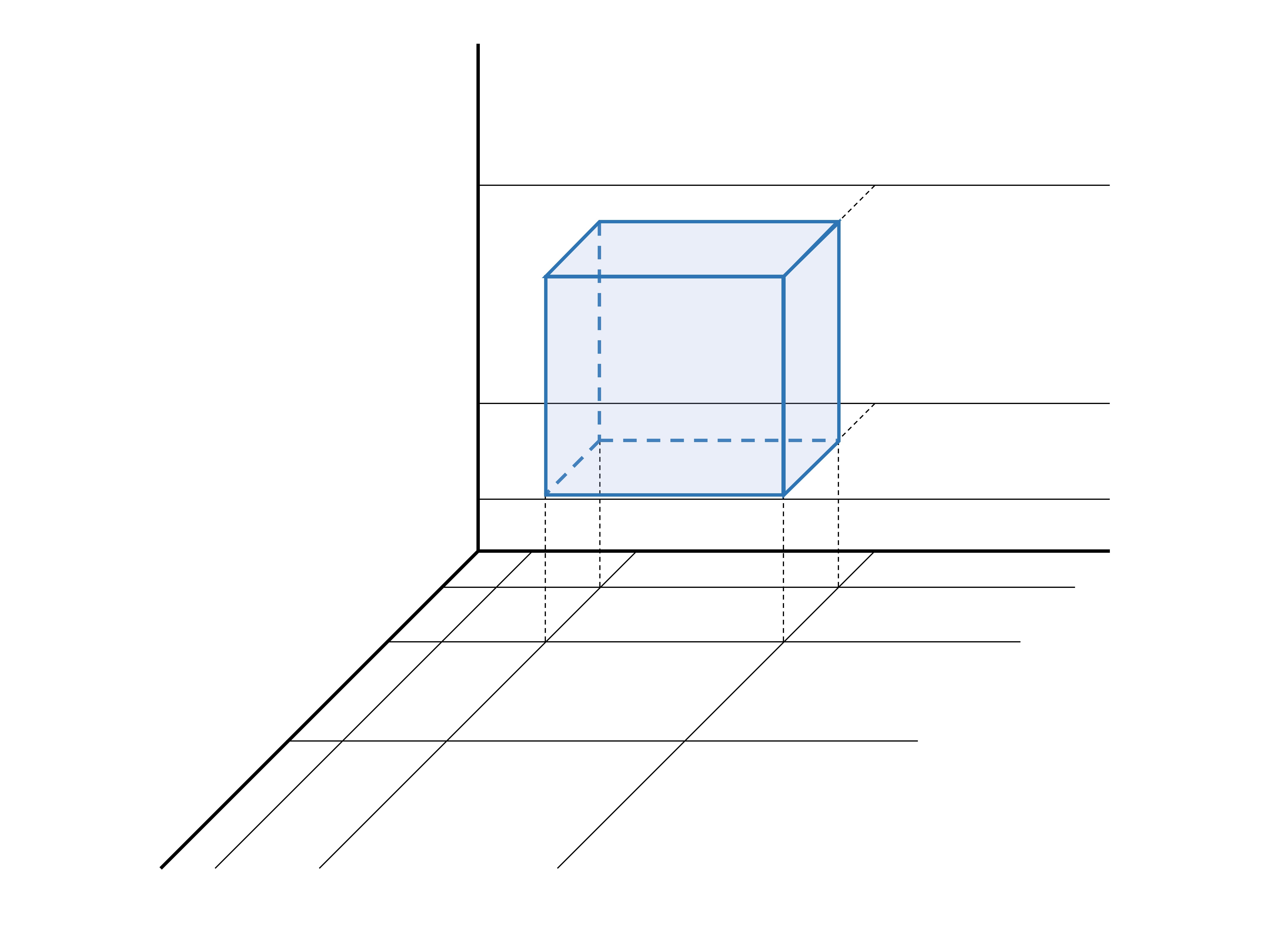}
    \caption{An example of the box used by the HCM algorithm in 3-dimensional coordinates.}
    \label{fig:5}
\end{figure}

However, it is impossible to select zones corresponding to K-complexes based solely on the results of t-SNE due to how it operates. Thus, we must look for K-complexes in the raw harmonic space. The HCM algorithm first splits the harmonic space into boxes. The dimensions of the boxes vary with distance from coordinate zero according to logarithmic scale (Figure~\ref{fig:5}).

During calibration the algorithm is supplied with a sample EEG record, annotated by a somnologist. It selects and stores the coordinates of the boxes that contain at least one point corresponding to a window centered over the zero crossing of a K-complex, as labelled by a somnologist. Afterwards, when the algorithm is used to detect K-complexes in a new record, it checks whether the coordinates of the points corresponding to the windows are situated within one of the stored boxes. If not, the window is left unmarked, otherwise the algorithm moves to the next step.

If the window falls within one of the saved boxes, the algorithm checks whether at least one of the window’s harmonics satisfies the threshold condition outlined in the section devoted to the KBP method, with \emph{p\textsubscript{t}}~=~0.80. This is done because the amplitude of a K-complex is noticeably greater than that of the general EEG signal, so that the power of at least one of its harmonics should be above the power of the same harmonic in the remaining 80\% of the windows.

\section{Results}\label{results}

\subsection{KBP algorithm}\label{res_kbp_algorithm}

As was described above, the KBP algorithm has six tuning values: \emph{p\textsubscript{t, \emph{P\textsubscript{1-4}}}}, \emph{p\textsubscript{t, \emph{S}}} and \emph{V\textsubscript{t}}. While theoretically it is possible to tune the algorithm parameters by hand, in practice, algorithm tuning is best done automatically, since K-complex detection quality is easily quantifiable. In case of the KBP algorithm, tuning was performed with a brute force approach by iterating through every combination of parameters within ranges that would be reasonable. The optimum combination was selected based on two conditions:

\begin{enumerate}
    \item TPR $\geq$ 99\%,
    \item Maximum PPV.
\end{enumerate}

This combination of tuning values and the results obtained with them are shown on Table~\ref{tab:1}.

\begin{table}
    \caption{KBP tuning results.}
    \label{tab:1}
    \centering
    \begin{tabularx}{0.95\textwidth}{|X X|X X|X X|X X|X X|X X|X X|X X|}
        \hline
        \multicolumn{16}{|c|}{Parameters} \\
        \hline

        \multicolumn{2}{|c|}{Window} & \multicolumn{2}{c|}{Window} & \multicolumn{2}{c|}{\emph{p\textsubscript{t, \emph{P\textsubscript{1}}}}} & \multicolumn{2}{c|}{\emph{p\textsubscript{t, \emph{P\textsubscript{2}}}}} & \multicolumn{2}{c|}{\emph{p\textsubscript{t, \emph{P\textsubscript{3}}}}} & \multicolumn{2}{c|}{\emph{p\textsubscript{t, \emph{P\textsubscript{4}}}}} & \multicolumn{2}{c|}{\emph{p\textsubscript{t, \emph{S}}}} & \multicolumn{2}{c|}{\emph{V\textsubscript{t}}} \\
        \multicolumn{2}{|c|}{length, s} & \multicolumn{2}{c|}{step, s} & & & & & & & & & & & & \\
        \hline

        \multicolumn{2}{|c|}{1.024} & \multicolumn{2}{c|}{0.1024} & \multicolumn{2}{c|}{0.81} & \multicolumn{2}{c|}{0.86} & \multicolumn{2}{c|}{0.98} & \multicolumn{2}{c|}{0.89} & \multicolumn{2}{c|}{0.7} & \multicolumn{2}{c|}{2} \\
        \hline

        \multicolumn{16}{|c|}{Quality} \\
        \hline

        \multicolumn{5}{|c|}{TPR} & \multicolumn{6}{c|}{PPV} & \multicolumn{5}{c|}{FPR} \\
        \hline

        \multicolumn{5}{|c|}{99.0\%} & \multicolumn{6}{c|}{48.5\%} & \multicolumn{5}{c|}{6.3\%} \\
        \hline

    \end{tabularx}
\end{table}

\subsection{HCM algorithm}\label{res_hcm_algorithm}

Unlike the KBP algorithm, the HCM algorithm has only one tuning parameter \emph{p\textsubscript{t}}, making it significantly easier to tune. Thus, it was tuned manually, and the best result is presented in Table~\ref{tab:2}

\begin{table}
    \caption{HCM tuning results.}
    \label{tab:2}
    \centering
    \begin{tabularx}{0.95\textwidth}{X X X X X X}
        \hline
        \multicolumn{6}{|c|}{Parameters} \\
        \hline

        \multicolumn{2}{|c|}{Window length, s} & \multicolumn{2}{c|}{Window step, s} & \multicolumn{2}{c|}{\emph{p\textsubscript{t}}} \\
        \hline

        \multicolumn{2}{|c|}{1.024} & \multicolumn{2}{c|}{0.1024} & \multicolumn{2}{c|}{0.80} \\
        \hline

        \multicolumn{6}{|c|}{Quality} \\
        \hline

        \multicolumn{2}{|c|}{TPR} & \multicolumn{2}{c|}{PPV} & \multicolumn{2}{c|}{FPR} \\
        \hline

        \multicolumn{2}{|c|}{87.5\%} & \multicolumn{2}{c|}{26.1\%} & \multicolumn{2}{c|}{17.1\%} \\
        \hline
         & & & & & \\
    \end{tabularx}
\end{table}

\subsection{Discussion}\label{discussion}

It is clear from Tables~\ref{tab:1} and~\ref{tab:2} that of the two algorithms KBP offers higher sensitivity, as well as a significantly lower proportion of false positive registrations. Thus, it is the KBP algorithm that we will compare to the other algorithms presented in other publications. This comparison is presented in Table~\ref{tab:3}. Note that two sets of results are given for the KBP algorithm, each of which corresponds to slightly different sets of tuning parameters. This is done to give a better comparison to the algorithms that have TPR around 90\%.

Of all the algorithms presented in previous works, the algorithm proposed in \cite{alsalman} displays the highest quality of K-complex detection, however this was achieved by using a neural network. Unfortunately, it is impossible to fully compare the two algorithms since the authors of \cite{alsalman} did not provide enough information to calculate PPV. Still, results presented in Table~\ref{tab:3} clearly demonstrate that one needs not employ neural networks to achieve a similar, or perhaps even higher quality of K-complex detection. Such an algorithm would have two inherent advantages over a neural network:

\begin{enumerate}
    \item Easy adjustability;
    \item Lower hardware requirements.
\end{enumerate}

\begin{table}
    \caption{Results comparison between different articles.}
    \label{tab:3}
    \centering
    \begin{tabularx}{0.85\textwidth}{|X X X|c|c|c|c|c|c|c|}
        \hline
        & \multicolumn{2}{|c|}{Desired} & \multicolumn{7}{c|}{Algorithm}\\
        \cline{4 - 10}

        & \multicolumn{2}{|c|}{value} & \multicolumn{2}{c|}{KBP} & \cite{bankman} & \cite{kam} & \cite{yucelbas} & \cite{shete} & \cite{alsalman}\\
        \hline

        TPR & \multicolumn{2}{|c|}{100\%} & 99.0\% & 90.7\% & 90.0\% & 75.0\% & 93.9\% & 90.9\% & 97.0\% \\
        \hline

        PPV & \multicolumn{2}{|c|}{100\%} & 48.5\% & 65.3\% & - & - & 29.7\% & 71.4\% & - \\
        \hline

        FPR & \multicolumn{2}{|c|}{0\%} & 6.3\% & 3.0\% & 8.1\% & 1.9\% & 9.3\% & 33.1\% & 5.8\% \\
        \hline
    \end{tabularx}
\end{table}

The latter is demonstrated quite well by the KBP algorithm, where an EEG signal 1 hour long is processed in less than 2 seconds on an Intel i5-10210U CPU.

\begin{figure}[t]
    \centering
    \includegraphics[scale=0.4]{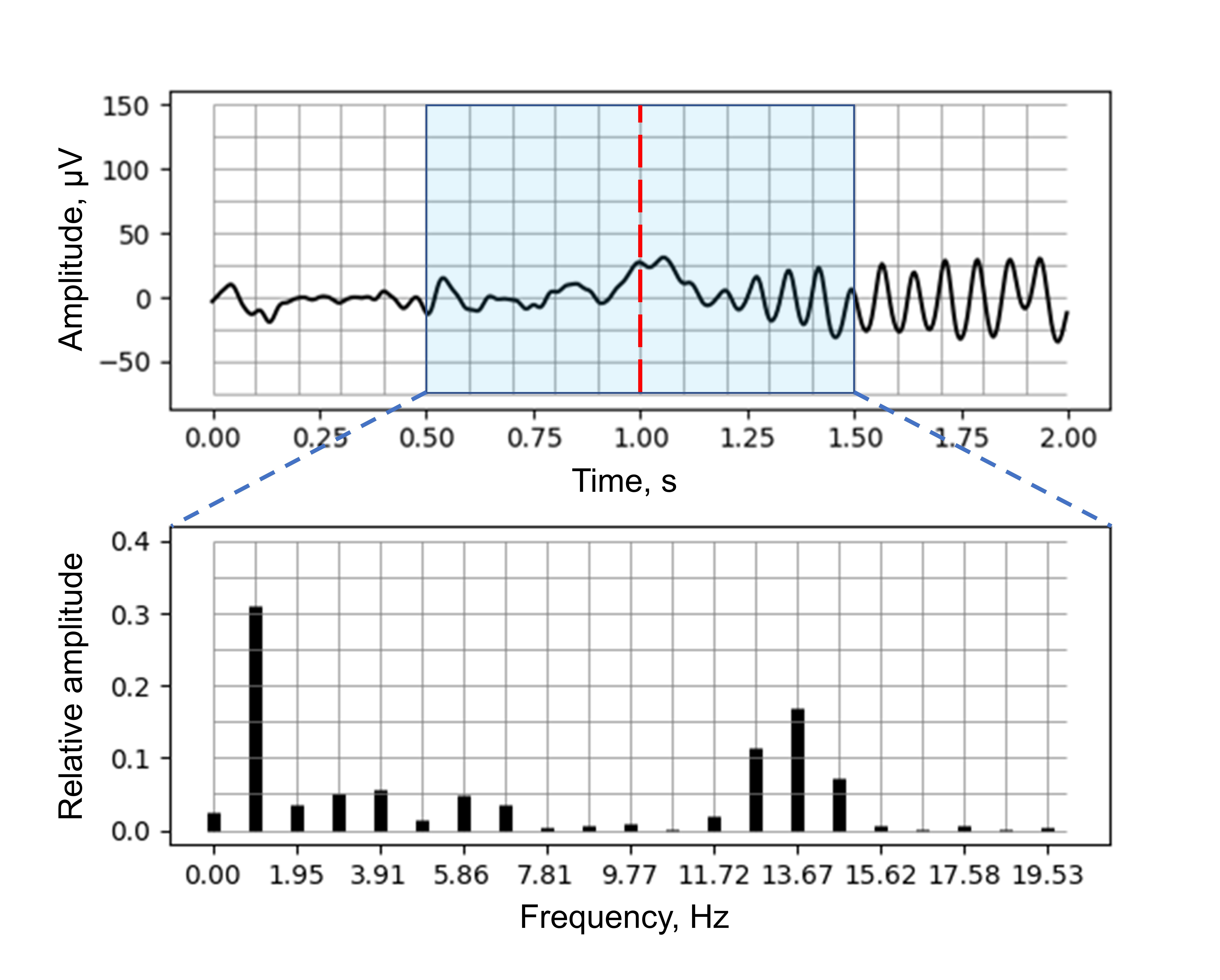}
    \caption{Unusual K-complex (red vertical line) and the spectrum within the window around it.}
    \label{fig:6}
\end{figure}

Nevertheless, there is still much room for improvement, specifically in terms of reducing the number of false positive registrations. One of the reasons for the PPV being so low is the lack of consistency in the way K-complexes are labeled by somnologists. For example, a detailed analysis of the annotations presented to us showed a significant number of signal fragments that were labelled as K-complexes but did not match the description provided in \cite{aasm}. An example of this is presented in Figure~\ref{fig:6}. 

Given the subjective way the K-complexes are labelled, it is also reasonable to assume, that the somnologists did not label every K-complex, which would also contribute to a decreased PPV. This is confirmed by the findings presented in \cite{devuyst}, where two somnologists presented annotations, that were significantly different.

\subsection{Conclusion}\label{conclusion}

The present article proposes two new mathematically simple methods of registering K-complexes in EEGs, which do not rely on neural networks and require little computing power. The quality of the proposed methods’ K-complex detection is estimated using both the traditional metric consisting of TPR and FPR, and the new proposed metric consisting of TPR and PPV, which is not dependent on the frequency of occurrence of the K-complexes in the EEG signal.

\subsection{Acknowledgements}\label{acknowledgements}

I would like to thank my supervisor, Vasiliy Dolmatov, and my colleague, Dr Igor Netay, for the valuable advice and for the many fruitful discussions we had on the subject of the present work.

\bibliographystyle{ieeetr}
\bibliography{ms}